\documentclass[10pt,twocolumn,letterpaper]{article}

\usepackage{iccv}
\usepackage{times}
\usepackage{epsfig}
\usepackage{graphicx}
\usepackage{amsmath}
\usepackage{amssymb}
\usepackage{booktabs}
\usepackage{algorithm}
\usepackage{algpseudocode}
\usepackage{amsmath}
\usepackage{graphics}
\usepackage{multirow}
\usepackage{multicol}
\usepackage{bm}
\usepackage{verbatim}
\usepackage{ulem}
\usepackage{float}
\usepackage{subfig}
\usepackage[marginal]{footmisc}

\usepackage[pagebackref,breaklinks=true,colorlinks,bookmarks=false]{hyperref}

\iccvfinalcopy 

\def\iccvPaperID{****} 

\ificcvfinal\fi

\begin{document}

\title{DiP: Learning Discriminative Implicit Parts for Person Re-Identification}

\author{Dengjie Li$^{*}$, Siyu Chen$^{*}$, Yujie Zhong, Lin Ma\\
Meituan\\
{\tt\small \{lidengjie,chensiyu25,zhongyujie\}@meituan.com}~~
{\tt\small forest.linma@gmail.com}
}
\maketitle

\footnote{~~~~$^{*}$Equal contribution.}
\ificcvfinal\thispagestyle{empty}\fi

\vspace{-0.4cm}
\begin{abstract}

In person re-identification (ReID) tasks, many works explore the learning of part features to improve the performance over global image features. Existing methods explicitly extract part features by either using a hand-designed image division or keypoints obtained with external visual systems. In this work, we propose to learn Discriminative implicit Parts (DiPs) which are decoupled from explicit body parts. Therefore, DiPs can learn to extract any discriminative features that can benefit in distinguishing identities, which is beyond predefined body parts (such as accessories). Moreover, we propose a novel implicit position to give a geometric interpretation for each DiP. The implicit position can also serve as a learning signal to encourage DiPs to be more position-equivariant with the identity in the image. 
Lastly, an additional DiP weighting is introduced to handle the invisible or occluded situation and further improve the feature representation of DiPs.
Extensive experiments show that the proposed method achieves state-of-the-art performance on multiple person ReID benchmarks.
\end{abstract}

\vspace{-0.2cm}
\section{Introduction}
\label{sec:intro}

Image-based person re-identification is an image retrieval problem. Given an image of a person captured from one camera, the main purpose of person ReID is to retrieve images of the same identity from another device. It may be difficult to distinguish a person with the same identity due to different camera views, occlusion, \textit{etc.}. At the same time, identities may be misidentified when they dress similarly. However, extracting global features cannot effectively solve the above problem. To alleviate this issue, lots of methods have been explored to extract more discriminative features through carefully designed discriminative part features. 

\begin{figure}
\setlength{\abovecaptionskip}{-0.2cm}
\begin{center}
\begin{tabular}{cccc}
    \includegraphics[width=0.2\linewidth]{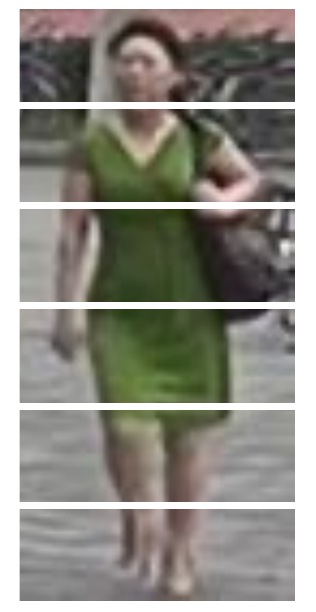} & 
    \includegraphics[width=0.2\linewidth]{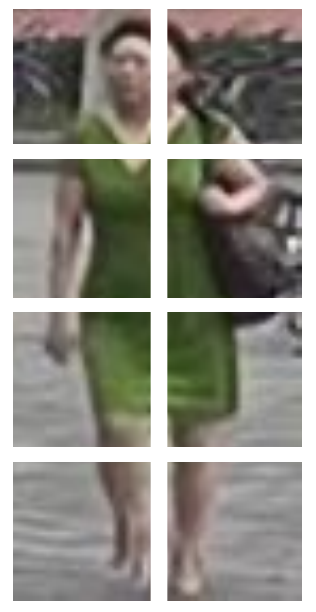} & 
    \includegraphics[width=0.2\linewidth]{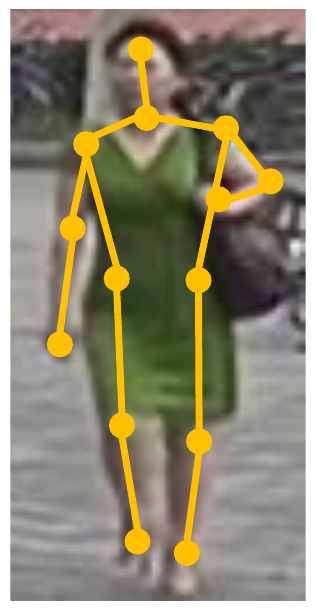} & 
    \includegraphics[width=0.2\linewidth]{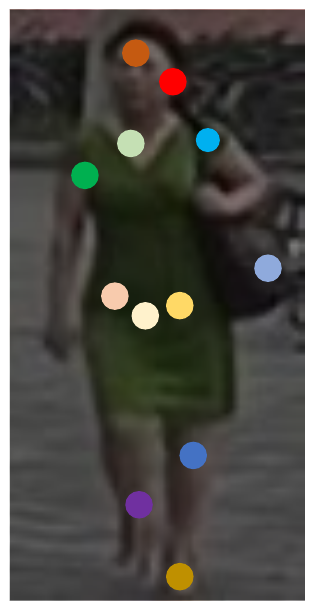} \\
    (a) & (b) & (c) & (d)
\end{tabular}
\end{center}
\caption{Comparison of part feature extraction for ReID. (a) Dividing the image horizontally\cite{sun2018beyond}; (b) patchify the image (\textit{e.g.} TransReID\cite{He_2021_ICCV}); (c) using an external system (\textit{e.g.} keypoints detection network\cite{zhao2017spindle}) to provide priors; (d) \textbf{ours}: the Discriminative implicit Parts (DiP) are learned end-to-end with the ReID network.}
\label{fig:motivation}
\end{figure}

\begin{figure}
\setlength{\abovecaptionskip}{-0.2cm}
\setlength{\belowcaptionskip}{-0.2cm}
\begin{center}
\begin{tabular}{cc}
    \includegraphics[width=0.45\linewidth]{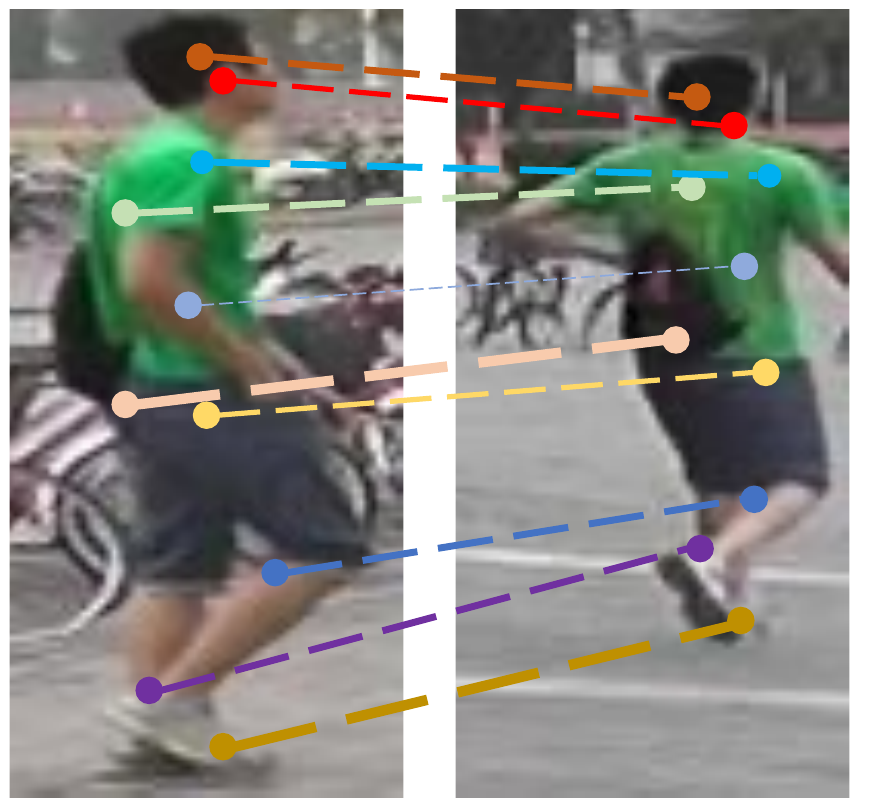} & 
    \includegraphics[width=0.45\linewidth]{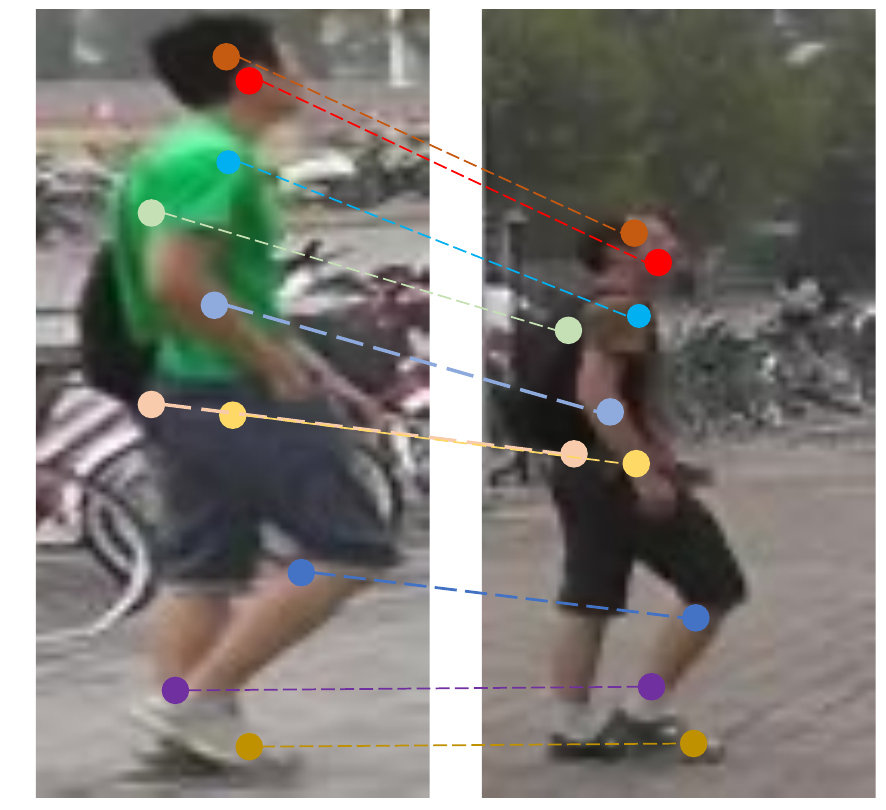} \\
    (a) & (b)
\end{tabular}
\end{center}
\caption{DiP enables a more fine-grained comparison between images. (a) illustrates the part matching between the same identity, while (b) is between different identities. The wider the dotted line, the closer the distance between the two parts in the feature space.}
\label{fig:part-based_dist}
\end{figure}

A group of methods is to split images horizontally to generate stripes (see Figure~\ref{fig:motivation}(a)), then extract the features of the stripes as part features for better matching \cite{sun2018beyond, wang2018learning}. Recently, TransReID\cite{He_2021_ICCV} follows the fashion of Vision Transformer\cite{dosovitskiy2020vit} to patchify the image and extract the patches' feature as local features (see Figure~\ref{fig:motivation}(b)). Broadly speaking, TransReID can be categorized into this group. This set of methods lacks flexibility and destroys the integrity of bodies. When there is heavy occlusion and pose variations, the bluntly generated horizontal stripes or image patches do not align the correct body parts well. 

For better alignment, another group of methods utilizes human parsing/keypoints detection/pose estimation to automatically generate different body parts (see Figure~\ref{fig:motivation}(c)) to fuse with global features \cite{suh2018part, zhao2017deeply, cheng2016person, zhao2017spindle, li2017learning}. 

Although this group of methods alleviates misalignment to a certain extent, they require additional skeletons to obtain priors. Moreover, not all discriminable regions are bound to explicit body parts. For example, accessories such as backpacks, hats, \textit{etc.}, although not belonging to body parts or key points, are still helpful for identification.

\vspace{0.2cm}
To decouple discriminative features from explicit body parts, we propose to learn \textit{Discriminative implicit Parts} (\textit{DiP}s for short) by extracting body-independent features that facilitate identity recognition. It makes our proposed method more concise than others, since it does not require any rigid image division rule, nor does it rely on any external priors (see Figure~\ref{fig:motivation}(d)). 

\vspace{0.2cm}
The extraction process of DiPs is simple. In a word, DiPs are extracted along with the backbone transformer network by appending randomly initialized tokens to the input sequence. We neither need to know the semantics of various body parts nor require each token to learn an explicit body part, that is, a DiP just represents an implicit part with discriminative features. Meanwhile, we construct an \textit{implicit position} for each DiP to give a geometric interpretation of the DiP in the image space. Additionally, we leverage the implicit position as a learning signal to make DiPs more position-equivariant with the person in the input image. Furthermore, we introduce the DiP weighting to better characterize each DiP in terms of its degree of visibility, which takes part in the proposed part-based distance computation (see Figure~\ref{fig:part-based_dist}). The part-based distance compares identities through fine-grained part features instead of the global feature, which further boosts the performance of our method.

\vspace{0.2cm}
Compared with previous works, our method can learn DiPs with small modifications to the network, which shows the efficiency and effectiveness of our method. The contributions of the paper are summarised below:

\vspace{-0.2cm}
\begin{itemize}
\setlength{\itemsep}{2pt}
\setlength{\parsep}{2pt}
\setlength{\parskip}{2pt}
\item We propose to learn \textit{Discriminative implicit Parts} (\textit{DiP}s) by extracting features that are beyond explicit body parts and can significantly improve the ReID performance.

\item We propose an \textit{implicit position} to give a geometric interpretation for each DiP in the image space. We also encourage the DiPs to learn to predict their implicit positions such that DiPs become position-equivariant with identities.

\item We further model the contribution of DiPs in images via the DiP weighting, and a part-based distance computation to enable a more fine-grained comparison between images.

\item DiP achieves state-of-the-art performance on multiple widely used benchmarks in person ReID, including  MSMT17\cite{wei2018person}, Market-1501\cite{zheng2015scalable}, Duke-ReID\cite{ristani2016performance}, CUHK03\cite{li2014deepreid}, and Occluded-Duke\cite{miao2019pose}, which demonstrates the superiority of our method.
\end{itemize}

\section{Related Work}
\label{sec:related}

\begin{figure*}
\setlength{\abovecaptionskip}{-0.2cm}
\begin{center}
    \includegraphics[width=0.95\linewidth]{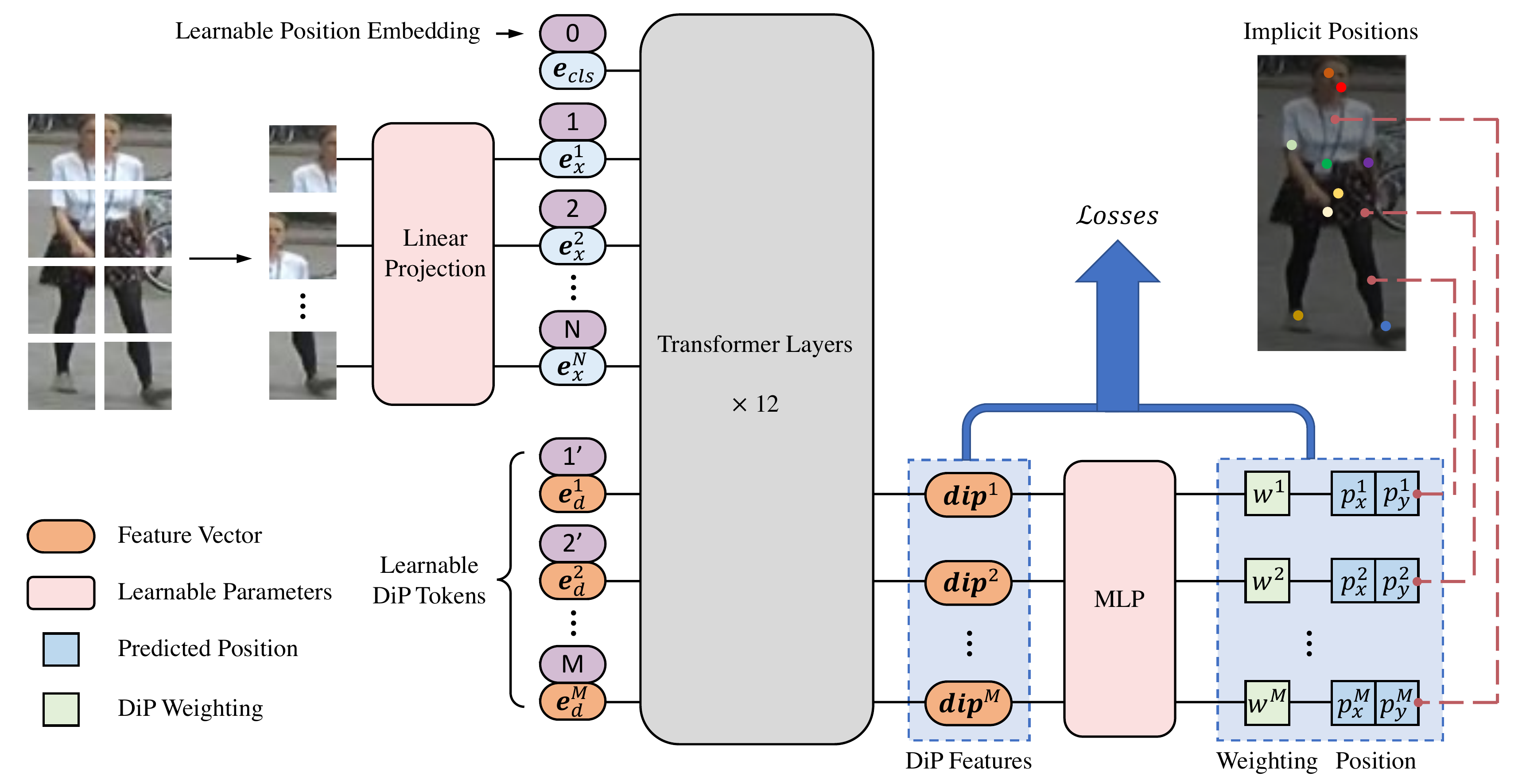}
\end{center}
   \caption{Overview of our method. The input is split into several image patches and linearly projected to token embeddings. We append a number of DiP tokens as Transformer input to extract DiP features. Each token adds a learnable position embedding to distinguish each other. An MLP is adopted to predict the position and DiP weighting. Each predicted position is supervised by the implicit position that is constructed via the correlations between the DiP and all patch features.}
\label{fig:overview}
\end{figure*}

\subsection{Part Feature Extraction}
\label{sec:part-level}
When occlusion and pose variations occur, global feature representation is not enough to provide fine-grained features for accurate person re-identification, thus many techniques turn to pursue fine-grained local/part features to obtain discriminative representations. Part feature representation learning methods can be classified into two types: the first one, by directly splitting the image horizontally, stripes can be obtained to serve as discriminative parts; the second one, additional skeletons such as pose estimation, keypoint detection, and human parsing are applied to automatically generate body parts. 

\vspace{-0.4cm}
\paragraph{Image division.}
PCB\cite{sun2018beyond} divides the image horizontally into six stripes and calculates the corresponding classification loss for each stripe. \cite{sun2019perceive} generates a part image by recombining stripes to learn the visibility-aware model. \cite{varior2016siamese} utilizes the LSTM network to sequentially process the stripes and fuses all image patches to obtain part features. \cite{zhang2017alignedreid} uses the shortest path distance to achieve automatic alignment of stripes. \cite{wang2018learning} introduces multiple granularities network to obtain discriminative features.

\vspace{-0.4cm}
\paragraph{Additional skeletons.}
\cite{zheng2019pose} uses the pose estimation network to estimate the keypoints of persons, then apply affine transformation to align the same body parts. \cite{zhao2017spindle} utilize keypoints to extract human body structure ROIs. To be more robust against pose changes, \cite{wei2017glad} proposed a global-local-alignment descriptor.

\subsection{Transformer-based ReID}
\label{sec:transformer-based}
With the development of Vision Transformer\cite{dosovitskiy2020vit} (ViT), some researchers have introduced it to person ReID task. TransReID\cite{He_2021_ICCV} is a pure transformer-based implementation that achieves state-of-the-art performance. It adds triple loss to ViT to build a strong baseline. In addition, they designed JPM (Jigsaw Patch Module) and SIE (Side Information Embeddings) to further boost the feature extraction process. The JPM module randomly re-groups the patch embeddings to form multiple branches. Each branch extracts the corresponding local features and is optimized the same as the global feature. The SIE module models non-visual information to help the network reduce the impact of the camera and perspective on feature extraction. 

NFormer\cite{wang2022nformer} uses transformer to model the relationship between image features to alleviate abnormal representations caused by high intra-identity variations. NFormer requires an extra network for feature extraction.

\section{Method}
\label{sec:method}

The overall architecture of the proposed method is illustrated in Figure~\ref{fig:overview}. It mainly consists of a backbone network and some additional components for extracting the \textit{Discriminative implicit Part} (\textit{DiP}) features.
The standard ID loss and triplet loss are adopted to optimize the whole network, similar to~\cite{chen2019abd, He_2021_ICCV, wang2018learning, wang2022nformer, chen2020cluster, chen2021video}.

\vspace{-0.4cm}
\paragraph{Backbone network.}
We use Vision Transformer\cite{dosovitskiy2020vit} (ViT) as our backbone for feature extraction. In vision transformer, given an input image $\textbf{\textit{X}} \in \mathbb{R}^{H \times W \times C}$, where $H, W, C$ represent the height, width, and channels of the image. The image is divided into a sequence of $N$ patches, where the patch resolution is $(P, P)$. Assuming that the dividing stride size is $S$, the patch number $N$ can be expressed as $N=N_H \times N_W=\left\lfloor\frac{H+S-P}{S}\right\rfloor \times\left\lfloor\frac{W+S-P}{S}\right\rfloor$\cite{He_2021_ICCV}. Each patch will be flattened into a 1D vector, and the $N$ flattened vectors are projected to $N$ token embeddings with $D$ dimensions by a linear projection layer. A [CLS] token is prepended to the token sequence, it acts as a global feature in the vanilla ViT\cite{dosovitskiy2020vit}. To model the spatial information of each patch, a learnable position embedding is added to each token.

\vspace{-0.4cm}
\paragraph{Part-based ReID.}
In our method, instead of learning global features, we pursue extracting DiPs for person re-identification. 
Each DiP is associated with a geometric interpretation which we term \textit{implicit position}. By learning to predict the implicit position, the DiPs become more position-equivariant with the identities.
In the following, we provide the details of DiPs (Section~\ref{sec:dip}), implicit position (Section~\ref{sec:position}), DiP weighting (Section~\ref{sec:weight}), and the inference stage (Section~\ref{sec:inference}).

\subsection{Discriminative Implicit Parts}
\label{sec:dip}
We posit that the [CLS] token is a global semantic representation, which lacks fine-grained discriminative part features. Although the patch token is an explicit image patch representation, it is not necessarily semantically meaningful. For example, the head is usually split into two different patches (see figure~\ref{fig:motivation}(b)). Therefore, we seek to learn meaningful part representations that are both locally discriminative and not limited by any hand-designed image division. We propose \textit{Discriminative implicit Parts} (\textit{DiP}s) to improve the method of recognizing identities by learning fine-grained discriminative features of the image. 
We consider the parts as \textit{implicit} because a DiP represents an identifiable area which can be a body part (such as the head or feet) or accessories (such as hats, backpacks), \textit{etc.}. DiPs are not limited to any explicit body parts but instead, focus on the parts that help distinguish identity from the others.

Figure~\ref{fig:overview} illustrates the DiPs extraction process. Similar to the [CLS] token in vanilla ViT, we simply add several part tokens as inputs to ViT, where each token also adds learnable positional embeddings to differentiate itself from others. The input sequence to the transformer layers can then be represented as:
\begin{equation}
\mathcal{Z}_0=\left[\textbf{\textit{e}}_{\mathrm{cls}} ; \textbf{\textit{e}}_x^1 ; \textbf{\textit{e}}_x^2 ; \cdots ; \textbf{\textit{e}}_x^N ; \textbf{\textit{e}}_d^1 ; \textbf{\textit{e}}_d^2 ; \cdots ; \textbf{\textit{e}}_d^M \right]+\mathcal{P},
\end{equation}
where $\textbf{\textit{e}}_{\mathrm{cls}}$ indicates [CLS] token; $\textbf{\textit{e}}_x^i$ indicates token embedding of $\textit{i}$-th patch in the sequence; $\textbf{\textit{e}}_d^i$ indicates $\textit{i}$-th part token that appended to learn discriminative implicit part; $N$ and $M$ are the number of patch tokens and part tokens, respectively. $\mathcal{P} \in \mathbb{R}^{(1+N+M)\times{D}}$indicates position embedding. $\mathcal{Z}_0$ indicates the input of transformer layers. Assuming that the encoder consists of $L$ layers, its output $\mathcal{Z}_L$ can be expressed as:
\begin{equation}
\mathcal{Z}_L=\left[\textbf{\textit{f}}_{\mathrm{cls}} ; \textbf{\textit{f}}^1 ; \textbf{\textit{f}}^2 ; \cdots ; \textbf{\textit{f}}^N ; \textbf{\textit{dip}}^1 ; \textbf{\textit{dip}}^2 ; \cdots ; \textbf{\textit{dip}}^M \right],
\end{equation}
where $\textbf{\textit{f}}^i$ indicates the $\textit{i}$-th patch feature that is extracted by transformer layers; $\textbf{\textit{dip}}^i$ is the $\textit{i}$-th feature that represents a discriminative implicit part of an image.

\subsection{Implicit Position}
\label{sec:position}

Ideally, DiPs represent discriminative parts after the learning process, and they should have the following two properties: (1) a DiP has a concrete geometric interpretation in the image space to represent the actual location of the part; (2) DiPs should be position-equivariant with the person in the input image, even if it encounters poses variations or occlusion. 
For the first property, we introduce the \textit{implicit position}, which is a coordinate in image space and computed based on the relation between each DiP and all the patch features.
To further enhance the second property, we propose to leverage the implicit position as a learning signal.
We elaborate on these two aspects in the following.

\begin{figure}
\setlength{\abovecaptionskip}{-0.2cm}
\begin{center}
    \includegraphics[width=1.0\linewidth]{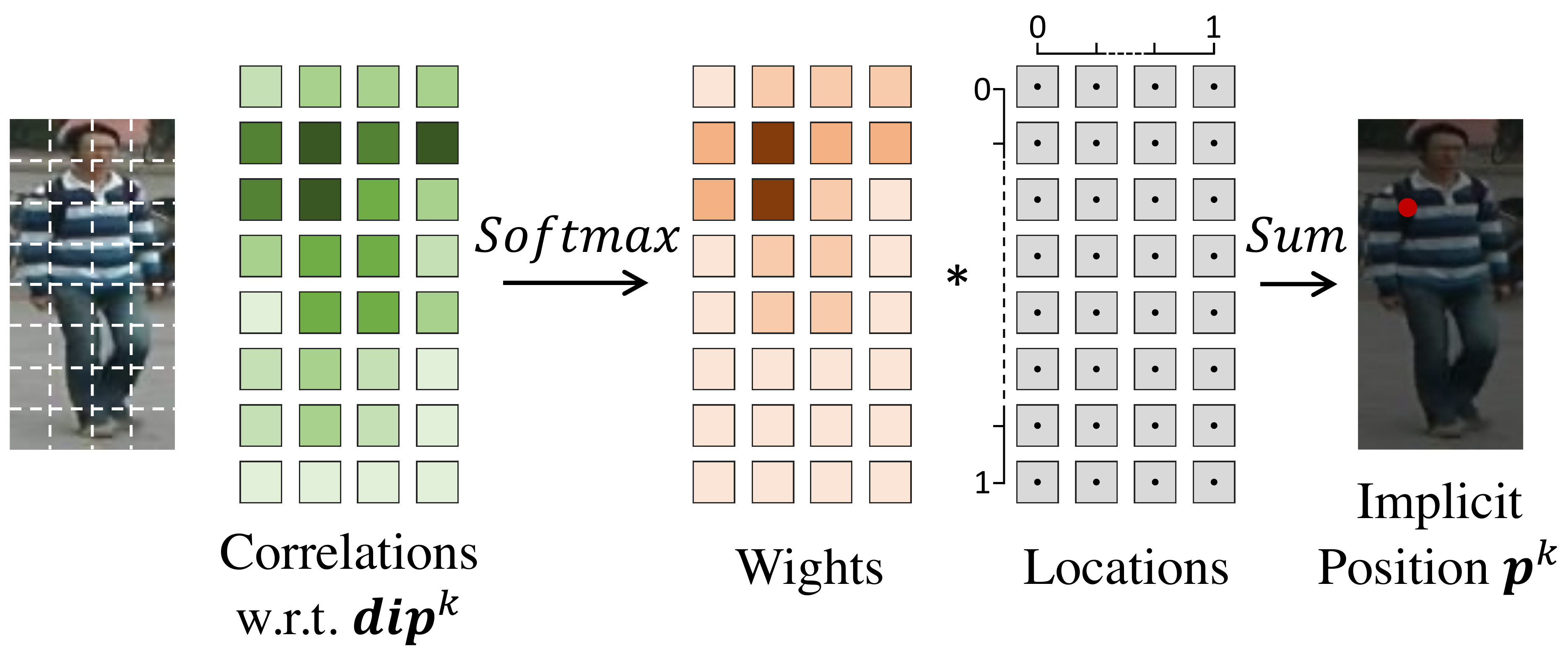}
\end{center}
   \caption{The details of implicit position construction. We calculate the similarity between $\textbf{\textit{dip}}^k$ and all the patch features to obtain correlation matrix $\textbf{\textit{C}}_k$, then we perform softmax to transfer correlations to weights. The implicit position is computed by weighted summation of the normalized location of each patch in image space.}
\label{fig:implicit-pos}
\end{figure}

\vspace{-0.4cm}
\paragraph{Implicit position.} 
We define the implicit position (\textit{i.e.} the coordinates in the image) of a DiP as the weighted sum of locations of the patches, where the weights are obtained based on the similarity between the DiP and patches in the feature space.
Figure~\ref{fig:implicit-pos} illustrates the process of obtaining the implicit position. Specifically, for an image with $N_H \times N_W$ patches, we normalize the location of the patch at the $(i,j)$ to be between $[0,1]$, that is, its normalized location $\textbf{\textit{l}}^{i,j}=(\frac{i}{N_H},\frac{j}{N_W})$. We compute the correlation matrix $\textbf{\textit{C}}_k$ with shape $N_H \times N_W$ for $\textbf{\textit{dip}}^k$, where each element $c^{i,j}_k$ is the inverse of the Euclidean distance between $\textbf{\textit{dip}}^k$ and patch feature $\textbf{\textit{f}}^{i,j}$. Then we perform softmax to all elements of $\textbf{\textit{C}}_k$ to get wight matrix $\textbf{\textit{W}}_k$. Thus, implicit position $\textbf{\textit{p}}^{k}=(p^k_{x}, p^k_{y})$ can be constructed by weighted-sum of patch locations:

\begin{equation}
p^k_x=\sum \limits_{i} \sum \limits_{j} \textbf{\textit{w}}^{i,j}*\textbf{\textit{l}}^{i,j}_x,~~~~
p^k_y=\sum \limits_{i} \sum \limits_{j} \textbf{\textit{w}}^{i,j}*\textbf{\textit{l}}^{i,j}_y.
\end{equation}

\vspace{-0.4cm}
\paragraph{Learning to predict implicit positions.} 
We encourage the network to correctly predict the implicit positions of DiPs in order to enhance the position-equivariance (with the person in the input image) of DiPs.
A simple MLP with 3 linear layers is adopted to make the prediction. 
Concretely, let $\hat{\textbf{\textit{p}}}=(\hat{p}_{x}, \hat{p}_{y})$ indicates the predicted position. The learning objective can be formulated as L2 regression, which we term as position-equivariance loss $\mathcal{L}_{PE}$:
\begin{equation}
\mathcal{L}_{PE}=\frac{1}{M} \sum \limits_{i=1}^M \left\|\textbf{\textit{p}}^{i}-\hat{\textbf{\textit{p}}}^{i}\right\|^2.
\end{equation}

\begin{figure}
\setlength{\abovecaptionskip}{-0.2cm}
\begin{center}
\begin{tabular}{cc}
    \includegraphics[width=0.45\linewidth]{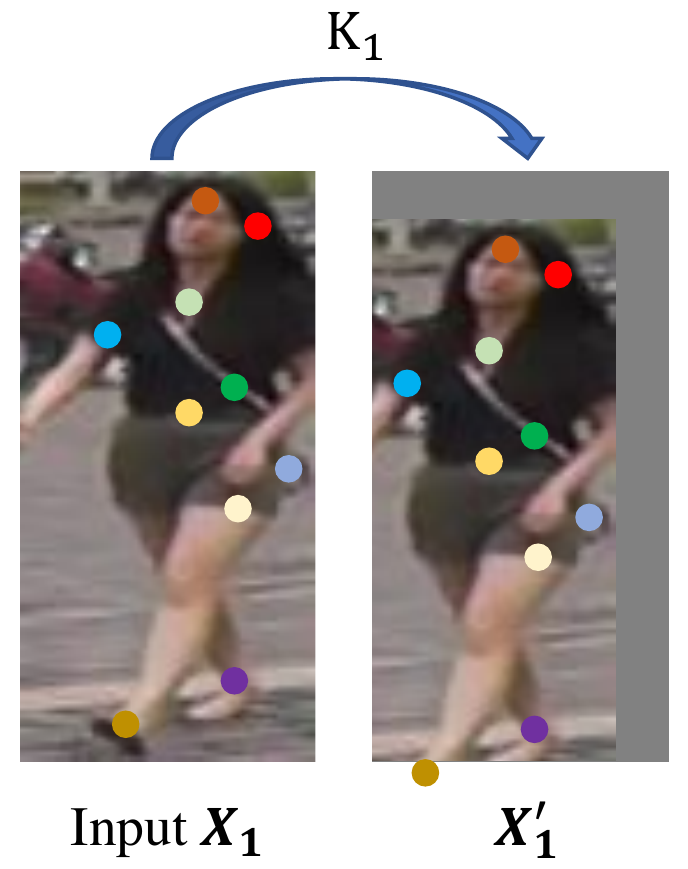} & 
    \includegraphics[width=0.45\linewidth]{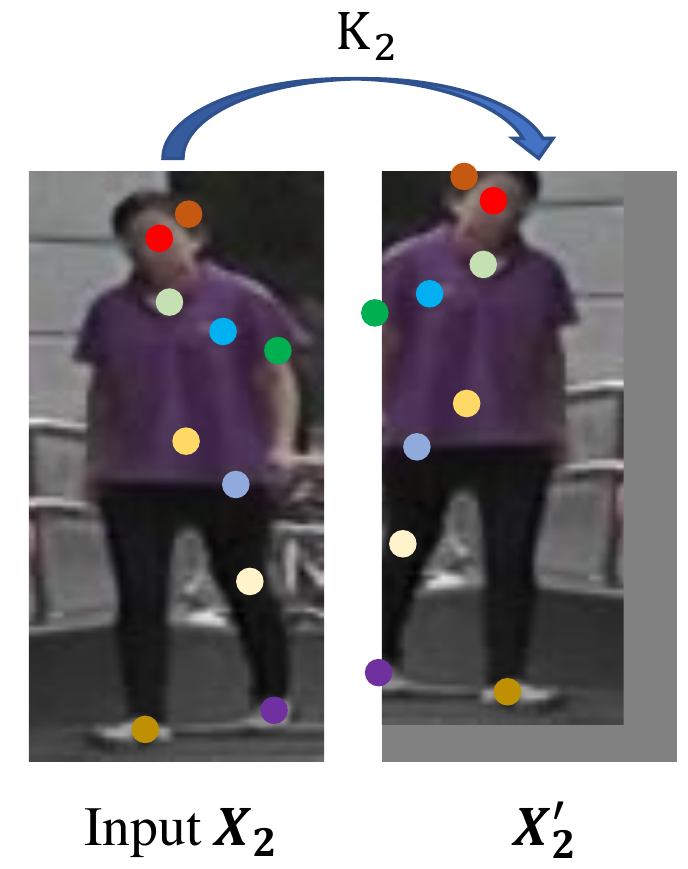} \\
    (a) & (b)
\end{tabular}
\end{center}
\caption{The transformed image and its corresponding implicit position are generated by applying affine transformation to the input image with a specific affine transformation matrix $\mathbf{K}$.}
\label{fig:affine_transformation}
\end{figure}

Further, to keep the predicted positions consistent for the same image, we utilize affine transformation to generate a transformed image of the input to let the prediction more transformation-invariance. We denote the input image as $\textbf{\textit{X}}$, then we apply affine transformation to $\textbf{\textit{X}}$ to generate transformed image $\textbf{\textit{X}}'$, which is used as another input to extract DiPs and predict the corresponding position. The transformation contains translation, scaling, and horizontal flipping (see Section~\ref{sec:implementation} for more details). Likewise, the implicit position that supervises the position prediction of $\textbf{\textit{X}}'$ is generated by applying affine transformation to implicit position $\textbf{\textit{p}}$ of the original image $\textbf{\textit{X}}$:
\begin{equation}
\left[\begin{array}{c}
p'_x \\
p'_y \\
1
\end{array}\right]=\underbrace{\left[\begin{array}{ccc}
a & b & c \\
d & e & f \\
0 & 0 & 1
\end{array}\right]}_{\mathbf{K}} \cdot
\left[\begin{array}{c}
p_x \\
p_y \\
1
\end{array}\right],
\end{equation}
where $(p_{x}, p_{y})$ and $(p'_{x}, p'_{y})$ denote the implicit position of image $\textbf{\textit{X}}$ and $\textbf{\textit{X}}'$, respectively; $\mathbf{K}$ denotes the affine transformation matrix. Figure~\ref{fig:affine_transformation} shows the transformed image and its corresponding implicit position. The position-equivariance loss for $\textbf{\textit{X}}'$ is formulated as:
\begin{equation}
\mathcal{L}'_{PE}=\frac{1}{M} \sum \limits_{i=1}^M \left\|{{\textbf{\textit{p}}'}}^{i}-{\hat{\textbf{\textit{p}}'}}^{i}\right\|^2,
\label{eq:align}
\end{equation}
where ${\hat{\textbf{\textit{p}}'}}^{i}=({{\hat{p'}}_{x}}^{i}, {{\hat{p'}}_{y}}^{i})$ denotes the predicted position of $\textbf{\textit{X}}'$. We point out that the transformed images also require computing ID loss and triplet loss. Notice that, image $\textbf{\textit{X}}'$ holds the same ID label with image $\textbf{\textit{X}}$.

\subsection{DiP Weighting}
\label{sec:weight}
In some images, one or more DiPs may be invisible or occluded. In this case, the invisible or occluded DiPs should have less or even zero contribution to the comparison between images.
Therefore, directly applying the part-based matching may lead to a distraction for the image-level ReID feature and the subsequent image matching. To alleviate the such problem, we introduce learnable weighting for each DiP.
Ideally, the weighting can dynamically adjust the contribution of the DiP when taking part in the loss computation at training and ReID distance calculation at inference. For each $\textbf{\textit{dip}}$, the weighting $w$ is a scalar with the value between (0,1) and predicted by the MLP introduced in section~\ref{sec:position}. 
Notably, all DiP weightings are learned (without labels) together with the network optimization process, participating in the computation of the ID loss\cite{zheng2017discriminatively} and the triplet loss\cite{liu2017end}, as discussed in the following.

\vspace{-0.4cm}
\paragraph{Role in ID loss.} For a specific image, we perform weighted summation to all $\textbf{\textit{dip}}$s with their corresponding DiP weighting $w$ to obtain mean feature $\widetilde{\textbf{\textit{dip}}}=\sum\limits^{M}_{i}w^i*\textbf{\textit{dip}}^i$. We use $\widetilde{\textbf{\textit{dip}}}$ as the feature representation of the image to optimize the ID loss, which is the cross-entropy loss in our implementation.

\vspace{-0.4cm}
\paragraph{Role in triplet loss.} We design a \textbf{part-based distance metric} to measure the distance between images to influence the hard sample mining process. Euclidean distance or cosine distance is usually used to measure the distance between two images. In our case, we first calculate the Euclidean distance between the corresponding DiPs of the two images $img_1$ and $img_2$. Let $\textbf{\textit{d}}$ indicates the obtained distance vector with $M$ elements, and $d^i \in \textbf{\textit{d}}$ is the $\textit{i}$-th element of $\textbf{\textit{d}}$ which indicates the distance between $\textbf{\textit{dip}}_{1}^i$ and $\textbf{\textit{dip}}_{2}^i$; $\textbf{\textit{w}}_1$ and $\textbf{\textit{w}}_2$ are weighting vectors with $M$ elements, representing the DiP weightings of $img_1$ and $img_2$, respectively. Then we compute the combined weightings which is denoted as $\textbf{\textit{w}}_c=\textbf{\textit{w}}_1*\textbf{\textit{w}}_2$. We apply softmax to the combined weightings $\textbf{\textit{w}}_c$ and perform weighted summation on $\textbf{\textit{d}}$ to obtain the final distance between the two images:
\begin{equation}
dist=\sum Softmax(\textbf{\textit{w}}_c)*\textbf{\textit{d}},
\end{equation}
where $*$ represents element-wise product; $Softmax($·$)$ indicates the softmax operation. We use this distance to pick triplet samples to calculate triplet loss $\mathcal{L}_{T}$.

\subsection{Inference}
\label{sec:inference}
At the inference stage, the predicted implicit position is not used and only the predicted DiP weightings participate in the \textbf{part-based distance calculation} (as described in Section~\ref{sec:weight}).

\section{Experiments}
\label{sec:experiments}

\begin{table*}
  \centering
  \begin{tabular}{@{}lccccccccccccc@{}}
    \toprule
    \multirow{2}*{Method} & \multirow{2}*{Size} & \multicolumn{2}{c}{MSMT17} & \multicolumn{2}{c}{Market-1501} & \multicolumn{2}{c}{Duke-reID} & \multicolumn{2}{c}{CUHK03-L} & \multicolumn{2}{c}{CUHK03-D} & \multicolumn{2}{c}{Occluded-Duke}\\
    ~ & ~ & R1 & mAP & R1 & mAP & R1 & mAP & R1 & mAP & R1 & mAP & R1 & mAP \\
    \midrule
    PCB+RPP\cite{sun2018beyond} & 256 & 68.2 & 40.4 & 93.8 & 81.6 & 83.3 & 69.2 & - & - & 63.7 & 57.5 & - & -\\
    MHN\cite{chen2019mixed} & 256 & - & - & 95.1 & 85.0 & 89.1 & 77.2 & 77.2 & 72.4 & 71.7 & 65.4 & - & -\\
    OSNet\cite{zhou2019omni} & 256 & 78.7 & 52.9 & 94.8 & 84.9 & 88.6 & 73.5 & - & - & 72.3 & 67.8 & - & -\\
    Pyramid\cite{zheng2019pyramidal} & 256 & - & - & 95.7 & 88.2 & 89.0 & 79.0 & 78.9 & 76.9 & 78.9 & 74.8 & - & -\\
    IANet\cite{hou2019interaction} & 256 & 75.5 & 46.8 & 94.4 & 83.1 & 87.1 & 73.4 & - & - & - & - & - & - \\
    STF\cite{luo2019spectral} & 256 & 73.6 & 47.6 & 93.4 & 82.7 & 86.9 & 73.2 & 68.2 & 62.4 & - & - & - & -\\
    BAT-net\cite{fang2019bilinear} & 256 & 79.5 & 56.8 & 94.1 & 85.5 & 87.7 & 77.3 & 78.6 & 76.1 & 76.2 & 73.2 & - & -\\
    PISNet\cite{zhao2020not} & 256 & - & - & 95.6 & 87.1 & 88.8 & 78.7 & - & - & - & - & - & - \\
    CBN\cite{zhuang2020rethinking} & 256 & - & - & 94.3 & 83.6 & 84.8 & 70.1 & - & - & - & - & - & - \\
    RGA-SC\cite{zhang2020relation} & 256 & 80.3 & 57.5 & \textbf{96.1} & 88.4 & - & - & 81.1 & 77.4 & 79.6 & 74.5  & - & -\\
    ISP\cite{zhu2020identity} & 256 & - & - & 95.3 & 88.6 & 89.6 & 80.0 & 76.5 & 74.1 & 75.2 & 71.4 & 62.8 & 52.3\\
    CBDB-Net\cite{tan2021incomplete} & 256 & - & - & 94.4 & 85.0 & 87.7 & 74.3 & 78.3 & 75.9 & 76.6 & 72.8 & 50.9 & 38.9\\
    CDNet\cite{li2021combined} & 256 & 78.9 & 54.7 & 95.1 & 86.0 & 88.6 & 76.8 & - & - & - & - & - & - \\
    PAT\cite{li2021diverse} & 256 & - & - & 95.4 & 88.0 & 88.8 & 78.2 & - & - & - & - & 64.5 & 53.6\\
    C2F\cite{zhang2021coarse} & 256 & - & - & 94.8 & 87.7 & 87.4 & 74.9 & 80.6 & 79.3 & \uline{81.3} & \textbf{84.1} & - & -\\
    TransReID\cite{He_2021_ICCV} & 256 & 83.3 & 64.9 & 95.0 & 88.2 & 89.6 & 80.6 & - & - & - & - & 64.2 & 55.7\\
    TransReID*\cite{He_2021_ICCV} & 256 & \uline{85.3} & 67.4 & 95.2 & 88.9 & 90.7 & 82.0 & - & - & - & - & \uline{66.4} & \uline{59.2}\\
    NFormer\cite{wang2022nformer} & 256 & 80.8 & 62.2 & 95.7 & \textbf{93.0} & 90.6 & \textbf{85.7} & 80.6 & 79.1 & 79.0 & 76.4 & - & -\\
    \textbf{DiP} (ours) & 256 & 84.6 & \uline{67.5} & 95.7 & 90.3 & \uline{91.2} & 83.8 & \uline{82.7} & \uline{80.5} & 80.2 & 77.7 & \uline{66.4} & 59.1\\
    \textbf{DiP}* (ours) & 256 & \textbf{86.3} &\textbf{70.6} & \uline{95.8} & \uline{90.8} & \textbf{91.6} & \uline{84.6} & \textbf{87.0} & \textbf{85.7} & \textbf{85.4} & \uline{83.1} & \textbf{68.0} & \textbf{60.8}\\
    \midrule
    MGN\cite{wang2018learning} & 384 & 76.9 & 52.1 & \uline{95.7} & 86.9 & 88.7 & 78.4 & 68.0 & 67.4 & 66.8 & 66.0 & - & -\\
    ABDNet\cite{chen2019abd} & 384 & 82.3 & 60.8 & 95.6 & 88.3 & 89.0 & 78.6 & - & - & - & - & - & - \\
    SCSN\cite{chen2020salience} & 384 & 83.8 & 58.5 & \uline{95.7} & 88.5 & 91.0 & 79.0 & \uline{86.8} & \uline{84.0} & \uline{84.7} & \uline{81.0} & - & - \\
    TransReID\cite{He_2021_ICCV} & 384 & 84.6 & 66.6 & 95.0 & 88.8 & 90.4 & 81.8 & - & - & - & - & - & - \\
    TransReID*\cite{He_2021_ICCV} & 384 & 86.2 & 69.4 & 95.2 & 89.5 & 90.7 & 82.6 & - & - & - & - & - & - \\
    \textbf{DiP} (ours) & 384 & \uline{86.6} & \uline{70.6} & 95.2 & \textbf{90.6} & \uline{91.2} & \uline{84.5} & 84.4 & 82.2 & 82.1 & 79.5 & \uline{68.8} & \uline{61.3}\\
    \textbf{DiP}* (ours) & 384 & \textbf{87.3} & \textbf{71.8} & \textbf{95.8} & \uline{90.3} & \textbf{91.7} & \textbf{85.2} & \textbf{87.4} & \textbf{85.6} & \textbf{84.9} & \textbf{82.7} & \textbf{71.1} & \textbf{63.1}\\
    \bottomrule
  \end{tabular}
  \caption{Performance of methods on ReID benchmarks. The star * in the superscript means dividing patches with small stride ($S=12$). R1 indicates Rank-1 accuracy and mAP indicates mean Average Precision. The value marked in bold means the best performance in each column and the value marked by underline means the second-best performance.}
  \label{tab:performance}
\end{table*}

\subsection{Datasets}
\label{sec:datasets}
To compare our performance with other methods, we conduct experiments on five popular ReID benchmarks: MSMT17\cite{wei2018person}, Market-1501\cite{zheng2015scalable}, Duke-ReID\cite{ristani2016performance}, CUHK03\cite{li2014deepreid}, and Occluded-Duke\cite{miao2019pose}. MSMT17 is the largest one which contains 4,101 IDs and 126,441 images; Market-1501 contains 1,501 IDs and 32,668 images. There are two versions of the CUHK03, the manual Labeled version and the Detected version. CUHK03 contains 1,467 IDs and 14,097 images. Duke-ReID include 1,404 IDs and 36,441 images. The images of Occluded-Duke are selected from Duke-ReID, and its training/query/gallery set contains 9\%/100\%/10\%/ occluded images respectively.

\subsection{Implementation}
\label{sec:implementation}
We select ViT-B as the backbone with weights pre-trained on ImageNet-21K and then finetuned on ImageNet-1K. For basic settings, the input images are resized to 256$\times$128. The batch size is set to 128. We use SGD as the optimizer with momentum set to 0.9. The initial learning rate is set to 0.04. The total training process lasts for 120 epochs. We use cosine LR scheduler to decrease the learning rate with warm-up strategy\cite{he2016deep, fan2019spherereid} applies to the first 5 epochs. For data augmentation, we only apply random erase\cite{zhong2020random}. We use Nvidia Tesla V100 GPU for training and inference.

For other hyperparameters, the number of DiPs $M$ is set to 12 for image size with $256\times128$ and set to 16 for $384\times128$. We conduct experiments with patch stride $S$ set to 16 and 12 under different image sizes to test the impact of patch strides on performance. The transformation to generate transformed image $\textbf{\textit{X}}'$ contains translation, scaling, and horizontal flipping. The translation includes $x$ and $y$ directions, with the image center as the coordinate origin, and the range is $[-0.12,0.12]$ times the corresponding image side length; the range of scaling is $[0.9,1.1]$; the probability of horizontal flipping is $0.5$.

\subsection{Comparison with SOTA methods}
\label{sec:comparison}
Table~\ref{tab:performance} shows the performance of our proposed method and other state-of-the-arts on MSMT17, Market-1501, DukeMTMC-reID, CUHK03, and Occluded-Duke. We report Rank-1 accuracy (R1) and mean Average Precision (mAP)\cite{zheng2015scalable} as evaluation protocols for comparison. Our method achieves state-of-the-art performance on multiple benchmarks or competitive results compared to the previous SOTAs.

\vspace{-0.4cm}
\paragraph{Results on MSMT17.}
MSMT17 is the largest and one of the most difficult benchmarks in the person ReID community. As shown in Table~\ref{tab:performance}, DiP achieves the best R1 and mAP over all previous competitors. For $256\times128$ image size setting, we achieve 67.5/70.6 mAP and 84.6/86.3 R1 for patch stride set to 16/12, which outperforms the second best method TransReID\cite{He_2021_ICCV} for 2.6/3.2 mAP and 1.3/1.0 R1. Likewise, we achieve 70.6/71.8 mAP and 86.6/87.3 R1 with $384\times128$ image size setting, which outperforms TransReID for 4.0/2.4 mAP and 2.0/1.1 R1. Compared with the recent Nformer\cite{wang2022nformer}, DiP surpasses it by 5.3/8.4 mAP and 3.8/5.5 R1 with the patch stride set to 16/12, respectively. Such a boost demonstrates the superiority of DiP. On top of that, we further improve the results to 71.8 mAP and 87.3 R1 with $384\times128$ image size and patch stride set to 12.

\vspace{-0.4cm}
\paragraph{Results on Market-1501.}
DiP achieves 90.3/90.8 mAP and 95.7/95.8 R1 for $256\times128$ image size setting with patch stride set to 16/12, respectively. At larger image size $384\times128$, we achieve 90.6/90.3 mAP and 95.2/95.8 R1 with different patch strides, which again outperforms TransReID\cite{He_2021_ICCV}.
NFormer achieves stunning performance on Market-1501. It is more suitable for scenarios when there are a large number of individual images (\textit{e.g.} 25.7 images for per identity in Market-1501) since NFormer learns information from adjacent persons/images. DiP aims to learn discriminative features in the image space, thus DiP is more suitable for scenes such as occlusion and complex data (\textit{e.g.} Occluded-Duke, MSMT17). Besides, NFormer requires more training epochs (160 epochs in total) and it uses an additional center loss for optimization.

\vspace{-0.4cm}
\paragraph{Results on DukeMTMC-reID.}
For image size $256\times128$, we achieve 83.8/84.6 mAP and 91.2/91.6 R1 with 16/12 patch stride, which beats TransReID by 3.2/2.6 mAP and 1.6/0.9 R1 with the same image size and patch stride. With a larger image size, our results can be improved to 84.5/85.2 mAP and 91.2/91.7 R1, which is comparable with NFormer with fewer training epochs.

\vspace{-0.4cm}
\paragraph{Results on CUHK03.}
On the manually labeled version of CUHK03, DiP achieves 80.5/85.7 mAP and 82.7/87.0 R1 with different patch strides, respectively. For the larger image size, we achieve 82.2/85.6 mAP and 84.4/87.4 R1 with a 16/12 patch stride. On the detected version, DiP is slightly inferior to C2F\cite{zhang2021coarse} by 1.0 mAP. But we achieve 85.4 R1 which is higher than C2F by 4.1 R1. For the $384\times128$ image size, DiP achieves 79.5/82.7 mAP and 82.1/84.9 R1 with different patch strides.

\vspace{-0.4cm}
\paragraph{Results on Occluded-Duke.}
Occluded-Duke is a difficult benchmark since it mostly contains occlusion data. Many methods are not tested on it. DiP hits better results than all previous state-of-the-arts, which is 60.8 mAP and 68.0 R1 with $256\times128$ image size. The best setting is $384\times128$ with 12 patch stride, which achieves 63.1 mAP and 71.1 R1. The results illustrate the effectiveness of DiP in dealing with occlusion.

\begin{table}
  \centering
  \setlength{\tabcolsep}{6px}
  \begin{tabular}{@{}lccccccc@{}}
    \toprule
    \multirow{2}*{~} & \multirow{2}*{$\mathcal{L}_{T}$} & \multirow{2}*{$\mathcal{L}_{PE}$} & \multirow{2}*{$\textbf{\textit{X}}'$} & \multicolumn{2}{c}{MSMT17} & \multicolumn{2}{c}{Market-1501} \\
    ~ & ~ & ~ & ~ & R1 & mAP & R1 & mAP \\
    \midrule
    0 & ~ & ~ & ~ & 81.1 & 60.5 & 93.8 & 86.2 \\
    \midrule
    1 & $\checkmark$ & ~ & ~ & 82.2 & 64.1 & 94.7 & 88.2 \\
    2 & $\checkmark$ & $\checkmark$ & ~ & 83.6 & 65.7 & 95.2 & 88.3 \\
    3 & $\checkmark$ & $\checkmark$ & $\checkmark$ & \textbf{84.6} & \textbf{67.5} & \textbf{95.7} & \textbf{90.3} \\
    \bottomrule
  \end{tabular}
  \caption{Ablation study on the losses and transformed image. Row 0 indicates that only the ID loss and Euclidean-based triplet loss are used; $\mathcal{L}_{T}$ indicates the triplet loss with the proposed part-based distance; $\mathcal{L}_{PE}$ indicates the position-equivariance loss; $\textbf{\textit{X}}'$ means the use of the transformed image that generates by applying affine transformation to input during training.}
  \label{tab:ab-loss}
\end{table}

\begin{table}
  \centering
  \setlength{\tabcolsep}{4px}
  \begin{tabular}{@{}ccccccc@{}}
    \toprule
    \multirow{2}*{DiP number} & \multirow{2}*{DiP Weighting} & \multicolumn{2}{c}{MSMT17} & \multicolumn{2}{c}{Market-1501} \\
    ~ & ~ & R1 & mAP & R1 & mAP \\
    \midrule
    8 & ~ & 84.6 & 67.5 & 95.1 & 89.3 \\
    8 & $\checkmark$ & \textbf{84.9} & \textbf{67.7} & \textbf{95.4} & \textbf{89.4} \\
    \midrule
    12 & ~ & 83.9 & 65.7 & 95.3 & 89.6 \\
    12 & $\checkmark$ & \textbf{84.6} & \textbf{67.5} & \textbf{95.7} & \textbf{90.3} \\
    \bottomrule
  \end{tabular}
  \caption{The effectiveness of DiP weighting with different numbers of DiPs. The part-based distance is used when comparing identities.}
  \label{tab:ab-vi}
\end{table}

\begin{figure}
\setlength{\abovecaptionskip}{0.2cm}
\setlength{\belowcaptionskip}{-0.4cm}
\centering
\subfloat[MSMT17]{\includegraphics[width=0.5\linewidth]{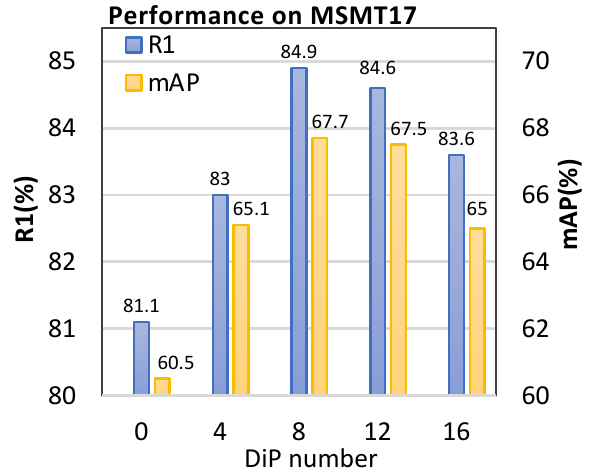}}
\subfloat[Market-1501]{\includegraphics[width=0.5\linewidth]{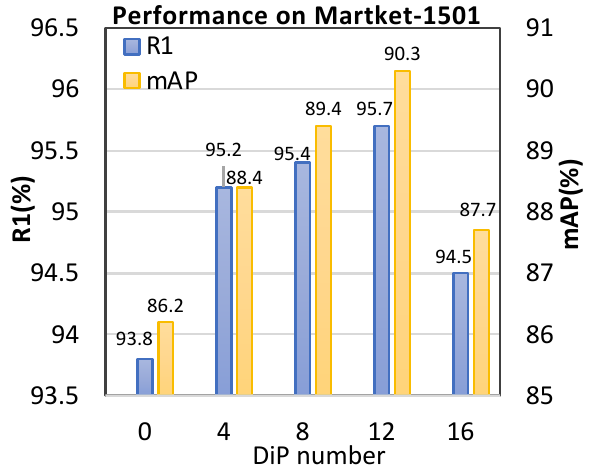}}

\caption{Ablation study of DiP number. The first column (number=0) reports the results of our baseline.}
\label{fig:ab-part}
\end{figure}

\begin{figure*}
\setlength{\abovecaptionskip}{0.1cm}
\setlength{\belowcaptionskip}{-0.2cm}
\centering
\subfloat[MSMT17]{\includegraphics[width=0.85\linewidth]{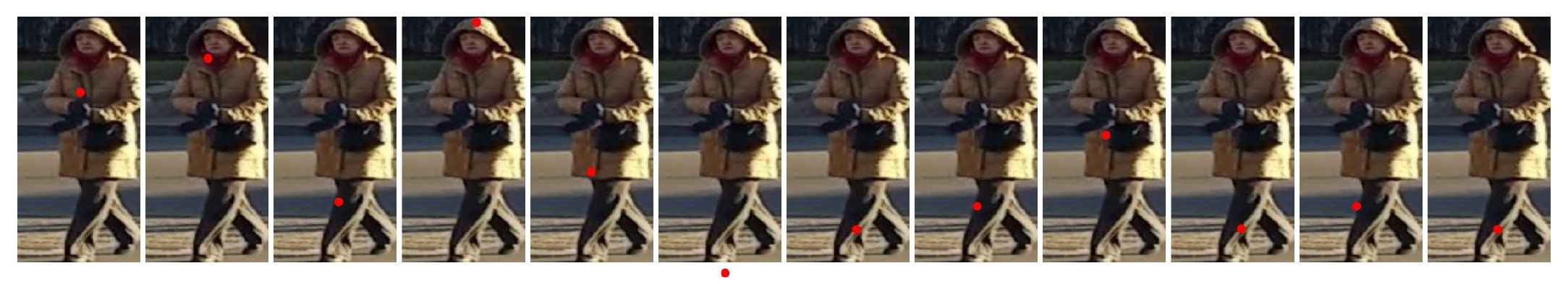}} \\
\subfloat[Market-1501]{\includegraphics[width=0.85\linewidth]{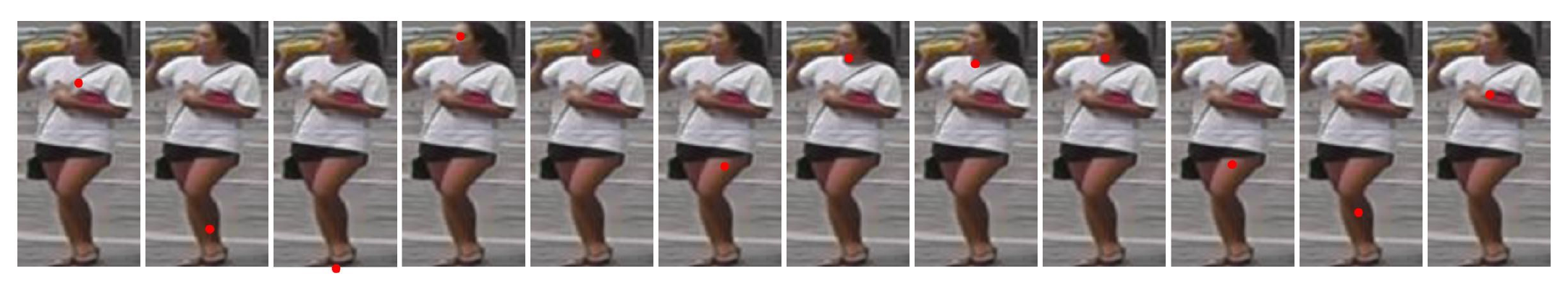}} 

\caption{Visualization of implicit positions. The title of each sub-figure indicates the dataset from which the identity comes. The dots in red indicate the predicted position of each DiP.}
\label{fig:vis}
\end{figure*}

\vspace{-0.4cm}
\subsection{Ablation Study}
\label{sec:ablation}
To demonstrate the effectiveness of our proposed components, we conduct comprehensive ablations for analysis. In this section, we ablate the proposed loss functions, the input image transformations, the DiP weighting, and the number of DiPs, respectively. All the experiments are studied on MSMT17 and Market-1501.

\vspace{-0.4cm}
\paragraph{Ablation on losses and transformed image.}
Table~\ref{tab:ab-loss} shows the results with different loss combinations. Row 0 is our baseline which only uses ID loss and Euclidean-based triplet loss for optimization, which achieves 81.1 R1/60.5 mAP on MSMT17 and 93.8 R1/86.2 mAP on Market-1501. When we replace Euclidean-based triplet loss with the proposed part-based triplet loss (row 1), both MSMT17 and Market-1501 results improve significantly. 
When $\mathcal{L}_{PE}$ is added, the performance on MSMT17 is further improved to 83.6 R1/65.7 mAP, and the results on Market-1501 being slightly improved, which may be due to the dataset being small that lacks complexity. The results on MSMT17 and Market-1501 are also obviously improved when applying the transformed image.

\vspace{-0.4cm}
\paragraph{Ablation of DiP weighting.}
The impact of DiP weighting on performance is shown in table~\ref{tab:ab-vi}. For the setting with the DiP number set to 8, the performance slightly drops by 0.3 R1/0.2 mAP on MSMT17 and 0.3 R1/0.1 mAP on Market-1501 when we remove the DiP weighting.
For the setting with the DiP number set to 12, the results drop by 0.7 R1/1.8 mAP on MSMT17 and 0.4 R1/0.7 mAP on Market-1501 without DiP weighting.

\vspace{-0.4cm}
\paragraph{Ablation of the number of DiPs.}
In Figure~\ref{fig:ab-part}, we analyze the effect of different numbers of DiPs on the results. The first column (number=0) indicates the baseline shown in Table~\ref{tab:ab-loss}. When setting the number from 0 to 4, a significant improvement can be seen on both MSMT17 and Market-1501, indicating the effectiveness of DiP. When we set the number to 8, better performance can be achieved on MSMT17, but all our reported $256\times128$ results are set the number to 12 for generalization consideration. The results on both MSMT17 and Market-1501 degrade when the number is set to 16. We conjecture that too many DiPs may cause the risk of overfitting and thus be detrimental to the performance.

\subsection{Visualization}
\label{sec:vis}
Figure~\ref{fig:vis} shows the visualization of the implicit positions of DiPs. In addition to focusing on body parts (such as the head, torso, legs, and feet), DiP also focuses on distinguishable parts according to the specific identity. In Figure~\ref{fig:vis}(a), a DiP corresponds to the scarf on the person. In Figure~\ref{fig:vis}(b), a DiP pays attention to the strap of the shoulder bag and what the person is carrying (which is probably a book).

\begin{figure}
\setlength{\abovecaptionskip}{0.1cm}
\setlength{\belowcaptionskip}{-0.5cm}
\centering
\subfloat[]{\includegraphics[width=0.25\linewidth]{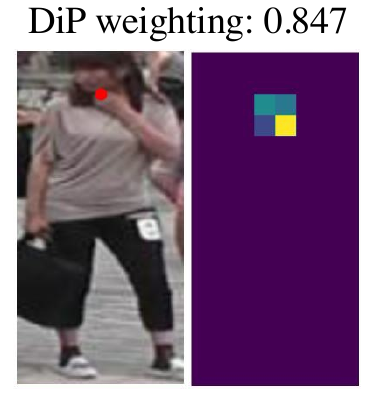}} 
\subfloat[]{\includegraphics[width=0.25\linewidth]{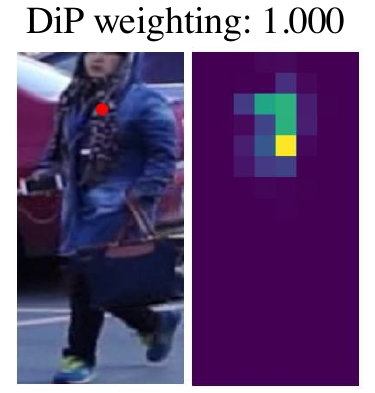}} 
\subfloat[]{\includegraphics[width=0.25\linewidth]{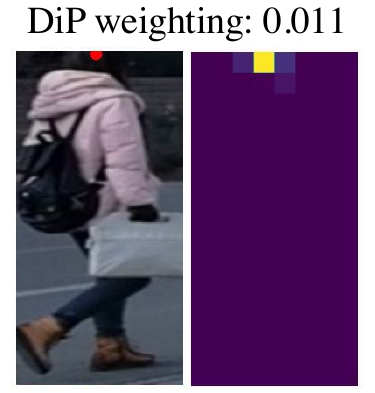}} 
\subfloat[]{\includegraphics[width=0.25\linewidth]{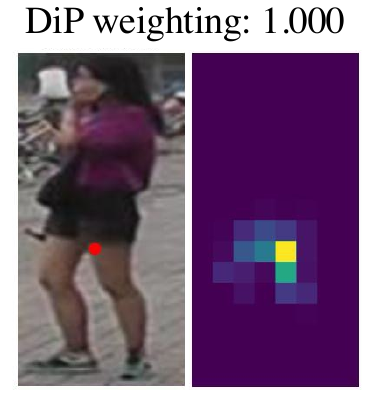}} 

\caption{Visualization of DiP weightings and score maps of different individuals. The red dot in each image denotes the implicit position, the score map on the right represents the corresponding weight matrix $\textbf{\textit{W}}_k$ (Section~\ref{sec:position}), and the value above represents its corresponding DiP weighting. Zoom in for more details.}
\label{fig:dipvis}
\end{figure}

Figure~\ref{fig:dipvis} shows the visualization of DiP weightings and score maps. 
From the score map in Figure~\ref{fig:dipvis}(a), it can be inferred that the DiP focuses on the neck area which is slightly occluded by the hand, thus it has a slightly lower weighting of 0.847.
The DiP in Figure~\ref{fig:dipvis}(b) focuses on the scarf that has no occlusion, therefore the DiP weighting is 1.0. The DiP in Figure~\ref{fig:dipvis}(c) focuses on the head with a very low DiP weighting, which may be due to the fact that part of the head is almost invisible. The DiP in Figure~\ref{fig:dipvis}(d) focuses on the edge of the shorts which is important for identifying the person, therefore it has a weighting of $1.0$.

\section{Conclusion}
\label{sec:conclusion}
In this paper, we propose a simple but effective method to learn discriminative implicit parts, which extracts features that are beyond explicit body parts to improve the performance of person re-identification. We introduce the implicit position to give a geometric interpretation for each DiP. We leverage the implicit position as a learning signal to promote the DiPs more position-equivariant with the input image. We further introduce to model the weighting of DiP and design a part-based distance computation to enable a more fine-grained comparison between images.
Our proposed method achieves state-of-the-art on multiple widely used person ReID benchmarks, which demonstrates the effectiveness of DiP.

\normalem
{\small
\bibliographystyle{ieee_fullname}
\bibliography{egbib}
}

\end{document}